\newif\ifabstract
\abstracttrue
 \abstractfalse 
\newif\iffull
\ifabstract \fullfalse \else \fulltrue \fi

\documentclass[11pt]{article}
\usepackage[utf8]{inputenc}

\usepackage{natbib}
\usepackage{graphicx}
\usepackage{authblk}

\usepackage{subcaption}
\usepackage{amsmath}
\usepackage{algorithm}
\usepackage[algo2e]{algorithm2e}
\usepackage{amssymb}
\usepackage{color}
\usepackage{epstopdf}
\usepackage{natbib}

\newtheorem{definition}{Definition}

\advance \oddsidemargin by -0.8in
\newcommand{\myparskip}{3pt}
\parskip \myparskip

\title{Scaling-up Split-Merge MCMC with Locality \\ Sensitive Sampling (LSS)}
\author{Chen Luo, Anshumali Shrivastava\thanks{Department of Computer Science, Rice University. Email: {\tt \{cl67, anshumali\}@rice.edu}. }}

\begin{document}

\maketitle

\begin{abstract}
Split-Merge MCMC (Monte Carlo Markov Chain) is one of the essential and popular variants of MCMC for problems when an MCMC state consists of an unknown number of components. It is well known that state-of-the-art methods for split-merge MCMC do not scale well. Strategies for rapid mixing requires smart and informative proposals to reduce the rejection rate. However, all known smart proposals involve expensive operations to suggest informative transitions. As a result, the cost of each iteration is prohibitive for massive scale datasets. It is further known that uninformative but computationally efficient proposals, such as random split-merge, leads to extremely slow convergence. This tradeoff between mixing time and per update cost seems hard to get around.
		
In this paper, we show a sweet spot. We leverage some unique properties of weighted MinHash, which is a popular LSH, to design a novel class of split-merge proposals which are significantly more informative than random sampling but at the same time efficient to compute. Overall, we obtain a superior tradeoff between convergence and per update cost. As a direct consequence, our proposals are around 6X faster than the state-of-the-art sampling methods on two large real datasets KDDCUP and PubMed with several millions of entities and thousands of clusters.
\end{abstract}

\section{Introduction}

Bayesian mixture models are of great interest due to their flexibility in fitting a countably infinite number of components which can grow with the data~\cite{medvedovic2004bayesian}.
	The growth of model complexity with the data is also in agreement with modern progress in machine learning over massive datasets. However, the appealing properties of Bayesian modeling come with hard computational problems. Even with simple mixture models, the mathematical problems associated with training and inference are intractable.
	As a result, recent research focuses on developing tractable computational techniques.
	In particular, the use of Markov chain Monte Carlo (MCMC) methods, to sample from the posterior distribution~\cite{andrieu2003introduction,nasrabadi2007pattern,wang2012split} is widely prevalent.
	The practical utility of these methods is illustrated in several applications including haplotype reconstruction~\cite{eronen2003markov},  nucleotide substitutions~\cite{huelsenbeck2001mrbayes}, gene expression~\cite{sharma2015computational}, etc.
	
	Metropolis-Hastings (MH) \cite{andrieu2003introduction} is a favorite class of MCMC methods, which includes several state-of-the-art algorithms that have proven useful in practice. MH is associated with a transition kernel which provides a proposal step. This step is followed by appropriate stochastic acceptance process that ensures detailed balance.
	A notable example of MH is the Split-Merge MCMC algorithm \cite{jain2004split, wang2015smart} which is particularly useful for problems where an MCMC state can be thought of as consisting of a number of components (or clusters). Here as the name suggests, the proposal step comprises of either a split or a merge. A split move partitions an existing mixture component (or cluster) into two, while a merge move combines two mixture components into one.
	
	In the seminal work of \cite{jain2004split}, split-merge MCMC procedure was proposed. To illustrate the process, the authors first introduce a random split-merge MCMC, where the split and the merge decision were taken uniformly at random. However, it was also pointed out, in the same paper, that due to the random nature of the proposal it was unlikely to lead to a new state $x'$ with higher likelihood $\mathcal{L}(x')$ leading to low acceptance. To mitigate the slow progress, the authors then propose the restricted Gibbs split-merge (RGSM). In RGSM instead of a random proposal the idea was to use restricted Gibbs sampling to generate proposals with a higher likelihood of acceptance. Thus, a less number of MCMC iterations were sufficient for convergence due to fewer rejections. However, the cost of restricted Gibbs is very high. As a result, even though the iterations are less, each iteration is costly making the overall algorithm slow, especially for large datasets. Our experiments confirm this slow convergence of RGSM.
	
	An essential and surprising observation about space asymmetry with smart proposals in split-merge MCMC was made in~\cite{wang2015smart}.
	The authors show the necessity to mix smart and dumb (random) proposals for faster progress. They proposed a Smart-Dumb/Dumb-Smart Algorithm (SDDS) as an alternative to RGSM. Instead of relying on Gibbs sampling, the SDDS algorithm uses the likelihood of the model itself as a guiding strategy for smart proposals. In other words, the SDDS method evaluates a large number of possible proposals $x'$ based on the likelihood of each $x'$ and choose the best ones. This strategy, as expected, ensures a higher chance of improving the state $x$ with every proposal. However, from a computational perspective, it is not difficult to see that smart proposal $x'$ obtained after evaluation of a large number of proposal states, based on the likelihood, is equivalent to evaluating all these states for acceptance/rejection as part of MH. As a result, the reduction in the number of iteration is not helpful in obtaining an efficient algorithm. Our experiments show that SDDS also has poor convergence.
	
	Unfortunately, most MCMC methodologies ignore the tradeoff between the number of iteration and computations associated with each iteration. They instead only focus on reducing the number of rejections, which is often achieved by informative proposals with increased per iteration cost. In this paper, we are interested in efficient split-merge MCMC algorithm which leads to overall fast convergence. Thus, reducing both is the aim of this work.
	
	{\bf Parallelization is Complementary:} Due to the significance of the problem there are several works which try to scale up MCMC by using parallelism. Parallelism is often achieved by running parallel MCMC chains on subsets of data and later merging them~\cite{chang2013parallel}. Since our proposal reduces the overall cost of split-merge MCMC algorithm in general, it will reduce the cost of each of the parallel chains thereby increasing the effectiveness of these parallelisms on MCMC. Thus, existing advances in parallelizing MCMC is complementary to our proposal.
	
	{\bf Our Contributions:}
	In this work, we show that it is possible to construct informative proposals without sacrificing the per-iteration cost. We leverage a simple observation that while designing proposals we can favor configurations where entities similar are likely to be in the same component. We use standard notions of vector similarity (cosine similarity) such as cosine or Weighted jaccard.
	To perform such sampling efficiently, we capitalize on the recent advances in LSH sampler~\cite{luo2017arrays,spring2017new,charikarhashing} that can perform adaptive sampling based on similarity. This forms our first proposal.
	
	Our first proposal leads to around 3x improvements over state-of-the-art methods. However, with similarity driven sampling, computing the Metropolis-Hastings (MH) ratio requires cost quadratic in the size of the cluster being split or merged. This is because while computing the state transition probability, we need to evaluate all possible ways that can lead to the desired split configuration. All these configurations have different probabilities due to similarity-based adaptive sampling and hence the probability computation is expensive. It appears at first that this cost is unavoidable. Surprisingly, it turns out that there is a rare sweet spot. With Weighted MinHash, we can design a split-merge proposal where the total cost of MH update is only linear in the size of the cluster being split or merged. The possibility is unique to MinHash due to its $k$-way generalized collision probability.  Our proposal and novel extension of MinHash collision probability could be of independent interest in itself.  To the best of our knowledge, it is not applicable to other LSH schemes.

	
	Overall, our proposed algorithms obtain a sweet tradeoff between the number of iteration and computational cost per iteration.  As a result, we reduce the overall convergence in time, not just in iterations.  On two large public datasets, our proposal MinHash Split-Merge (MinSM) significantly outperforms other state-of-the-art split-merge MCMC algorithms in convergence speed as measured on wall clock time on the same machine. Our proposed algorithm is around 6x faster than the second best baseline on real datasets without loss in accuracy.
	\section{Background}

	Our work requires bridging Locality Sensitive Sampling with split-merge MCMC algorithm. We briefly review the necessary background.
	\subsection{Locality Sensitive Hashing}
	\label{sec:srp}
	Locality-Sensitive Hashing (LSH) is a popular technique for efficient approximate nearest-neighbor search. LSH is a family of functions, such that a function uniformly sampled from this hash family has the property that, under the hash mapping, similar points have a high probability of having the same hash value.
	More precisely, consider $\mathcal{H}$ a family of hash functions mapping $\mathbb{R}^D$ to a discrete set $[0,R-1]$.
	\begin{definition} \label{def:lsh}{\bf Locality Sensitive Hashing (LSH) Family}\ A family $\mathcal{H}$ is called $(S_0,cS_0,u_1,u_2)$-sensitive if for any two points $x,y \in \mathbb{R}^d$  and $h$ chosen uniformly from $\mathcal{H}$ satisfies the following:
		\begin{itemize}
			\item if $Sim(x,y)\ge S_0$ then ${Pr}_\mathcal{H}(h(x) = h(y)) \ge u_1$
			\item if $ Sim(x,y)\le cS_0$ then ${Pr}_\mathcal{H}(h(x) = h(y)) \le u_2$
		\end{itemize}
	\end{definition}
	A collision occurs when the hash values for two data vectors are equal, meaning that $h(x) = h(y)$. LSH is a very well studied topic in computer science theory and database literature. There are many well-known LSH families in the literature. Please refer~\cite{gionis1999similarity} for details.
	\subsubsection{Locality Sensitive Sampling (LSS) and Unbiased Estimators}
	\label{sec:lshsample}
	LSH was considered as a black-box algorithm for similarity search and dimensionality reduction. Recent research~\cite{spring2017new} found that LSH can be used for something more subtler but useful. It is a data structure that can be used for efficient dynamically adaptive sampling.
	We first describe the sampling algorithm of~\cite{spring2017new} and later comment on its properties crucial to our proposal.
	
	The algorithm uses two parameters - $(K, L)$. We construct $L$ independent hash tables from the collection $\mathcal{C}$. Each hash table has a meta-hash function $H$ that is formed by concatenating $K$ random independent hash functions from some appropriate locality sensitive hash family $\mathcal{H}$. The candidate sampling algorithm works in two phases \cite{spring2017new}:
	\begin{enumerate}
		\item {\bf Pre-processing Phase:} We construct $L$ hash tables from the data by storing all elements $x \in \mathcal{C}$. We only store pointers to the vector in the hash tables because storing whole data vectors is very memory inefficient. This is one-time linear cost.
		\item {\bf Sampling Phase:} Given a query $q$, we collect one bucket from a randomly selected hash table and return a random element from the bucket. If the bucket is empty, we reselect a different hash table again. Keep track of the number of different tables $T$ probed.
	\end{enumerate}
	It is not difficult to show that an item returned as a candidate from a $(K, L)$-parameterized LSH algorithm is sampled with probability exactly  $1-(1-p^K)^L \times \frac{1}{Size}$, where $p$ is the collision probability of LSH function and $Size$ is the number of elements in the bucket.
	The LSH family defines the precise form of $p$ used to build the hash tables. Specifically, when $L=1$ and $K=1$, the probability reduced to the collision probability itself ($p$).
	Our proposal will heavily rely on this unusual probability expression $1-(1-p^K)^L \times \frac{1}{Size}$ to design an informative and proposal distribution.

	\subsection{Weighted (or Generalized) MinHash}
	\label{sec:wmin}
	
	Weighted Minwise Hashing is a known LSH for the Weighted Jaccard similarity~\citep{leskovec2014mining}. Given two positive vectors $x, \ y \in \mathbb{R^D}$, $x, \ y > 0$, the (generalized) Weighted Jaccard similarity is defined as $
	\mathbb{J}(x,y) = \frac{\sum_{i=1}^D\min\{x_i,y_i\}}{\sum_{i=1}^D\max\{x_i,y_i\}}.
	$, where $\mathbb{J}(x,y)$ is a frequently used measure for comparing web-documents~\citep{leskovec2014mining}, histograms (specially images), gene sequences, etc. 
	
	Weighted Minwise Hashing (WMH) (or Minwise Sampling) generates randomized hash (or fingerprint) $h(x)$, of the given data vector $x \ge 0$,  such that for any pair of vectors $x$ and $y$, the probability of hash collision (or agreement of hash values) is given by $Pr(h(x) = h(y)) = \frac{\sum \min\{x_i,y_i\}}{\sum \max\{x_i,y_i\}}$.
	
	A unique property of Minwise Hashing is that there is a natural extension of $k$-way collision~\cite{shrivastava2013beyond}.
	In particular, given vectors $x^{(1)}, \ x^{(2)}, ..., x^{(s)}$, the simultaneous collision probability is given by:
	\begin{equation}
	\begin{split}
	\label{eq:kwayCollProb}
	Pr(h(x^{(1)}) = h(x^{(2)}) = ... = h(x^{(s)})) = \frac{\sum_{j}^{D} \min\{x_j^{(1)}, x_j^{(2)},... ,x_j^{(s)}\}}{\sum_{j}^{D} \max\{x_j^{(1)}, x_j^{(2)},... ,x_j^{(s)}\}}
	\end{split}
	\end{equation}
	Minwise hashing can be extended to negative elements using simple feature transforms~\cite{li2017linearized}, which essentialy doubles the dimentions to 2D. In this paper, MinHash and Weighted MinHash denote the same thing.
	
	\subsection{Split-Merge MCMC}
	\label{sec:smmcmc}
	Split-Merge MCMC \cite{hughes2012effective} is useful for dealing with the tasks such as clustering or topic modeling where the number of clusters or components are not known in advance. Split-Merge MCMC is a Metropolis-Hastings algorithm with two main transitions: Split and Merge. During a split, a cluster is partitioned into two components. On the contrary, a merge takes two components and makes them to one.
	
	During the MCMC inference process, split and merge moves simultaneously change the number of entities and change the assignments of entities to different clusters.  \cite{jain2004split} proposes the first non-trivial Restricted Gibbs Split-Merge (RGSM) algorithm, which was later utilized for efficient topic modeling over large datasets in~\cite{wang2012split}.
	
	In \cite{wang2015smart}, the authors presented a surprising argument about information asymmetry. It was shown that both informative split and merge leads to poor acceptance ratio. The author proposed a  combination of the smart split with dumb (random) merge and dumb split with smart merge as a remedy.  The algorithm was named as Smart-Dumb/Dumb-Smart Split Merge algorithm (SDDS), which was superior to RGSM. To obtain non-trivial smart split (or merge), the authors propose to evaluate a large number of dumb proposals based on the likelihood and select the best. This search process made the proposal very expensive. It is not difficult to see that finding a smart split is computationally not very different from running a chain with several sequences of dumb (random) splits.
	\section{LSS based Split-Merge MCMC}
	\label{sec:lshsm}
	
	
	{\bf Utilizing Similarity Information:} In this paper, we make an argument that similarity information, such as cosine similarity, between different entities is almost always available. For example, in the clustering task, the vector representation of the data is usually easy to get for computing the likelihood. Even in an application where we deal with complex entities such as trees, it is not uncommon to have approximate embeddings~\cite{bengio2010label}.
	
	It is natural to believe that similar entities, in terms of cosine similarity or Jaccard distance, of the underlying vector representation, are more likely to go to the same cluster than non-similar ones. Thus, designing proposals which favor similar entities in the same cluster and dissimilar entities in different clusters is more likely to lead to acceptation than random proposals.
	
	However, the problem is far from being solved. Any similarity based sampling requires computing all pairwise similarity as a prerequisite, which is a quadratic operation $O(n^2)$. Quadratic operations are prohibitive (near-infeasible) for large datasets. One critical observation is that with the modern view of LSH as samplers, described preview section, we can get around this quadratic cost and design cheaper non-trivial proposals.
	
	\subsection{Naive LSS based Proposal Design}

	\label{sec:proposal}
	
	This section discusses how LSH can be used for efficient similarity sampling which will lead to an informative proposal. In addition, we also want the cost of computing the transition probabilities $q(x'|x)$, which is an important component of the acceptance ratio $\alpha(x'|x)$ \cite{jain2004split}, to be small. Here, $x$ denotes the state before split/merge, and $x'$ denote the state after split/merge.
	For a good proposal design, it is imperative that $q(x'|x)$ is easy to calculate as well as the proposed state $x'$ is informative. Thus, designing the right MCMC proposal process is the key to speed up computation. Following the intuition described before, we introduce our LSS based proposal design in the rest of this section.
	
	
	We first create the hash tables $T$ for sampling.
	We use Sign Random Projection as the LSH function, thus our notion of similarity is cosine.
	We pay a one-time linear cost for this preprocessing. Note, we need significantly less $K$ and $L$ (both has value 10 in our experiments) compared to what is required for near-neighbor queries as we are only sampling. The sampling is informative (better than random) for any values of $K$ and $L$
	
	For our informative proposal, we will need capabilities to do both similarity sampling as well as dissimilarity sampling for merge and split respectively. The similarity sampling is the usual sampling algorithm discussed in previous sections, which ensures that given a query $u$, points similar to $u$ (cosine similarity) are more likely to be sampled.
	Analogously, we also need to sample points that are likely to be dissimilar. With cosine similarity, flipping the sign of the query, i.e., changing $u$ to $-u$ will automatically do dissimilarity sampling.
	
	Inspired from~\cite{wang2015smart}, we also leverage the information asymmetry and mix smart and dumb moves for better convergence. However, this time our proposals will be super efficient. At each iteration of MCMC, we start by choosing randomly between an LSH Smart-split/Dumb-merge or an LSH Smart-merge/Dumb-split operation. These two operations are defined below:
	\subsubsection{Naive LSH Smart-split/Dumb-merge}
	LSH based split begins by randomly selecting an element $u$ in the dataset. Then, we use LSS (Locality-sensitive Sampler) to sample points likely to be dissimilar to $u$. Thus, we query our data structure $T$ with $-u$ as the query to get another element $v$ which is likely far away from $u$. If $u$ and $v$ belong to the same cluster $C$, we split the cluster. During the split, we create two new clusters $C_u$ and $C_v$. We assign $u$ to $C_u$ and $v$ to $C_v$. For every element in $C$, we randomly assign them to either $C_u$ or $C_v$. Since we ensure that dissimilar points $u$ and $v$ are split, this is an informative or smart split.
	If we find $u$ and $v$ are already in a different cluster, we do a dumb merge: randomly select two components, and merge these two components into one component.
	
	The most important part is that we can precisely compute the probability of the proposed split move $q(x'|x)$ and the corresponding inverse move probability $q(x|x')$ as follow:
	
	\begin{equation}
	\label{eq:lshsp}
	\begin{split}
	& q(x'|x) = \left(\frac{1}{2}\right)^{|C_u| + |C_v| - 2}
	 \sum_{u}^{C_u} \sum_{v}^{C_v} \left( \frac{1}{n} \left(1-\left(1-Pr(-u,v)^K\right)^L \right) \frac{|C_v \cap S_{-u}|}{|S_{-u}|} \right).
	\\& = 
	\frac{\sum_{u}^{C_u} \sum_{v}^{C_v} \left( \frac{1}{n} \left(1-\left(1-Pr(-u,v)^K\right)^L \right) \frac{|C_v \cap S_{-u}|}{|S_{-u}|} \right)}{2^{|C_u| + |C_v| - 2}}
	\\&
	q(x|x') = \frac{2}{M_{x'}(M_{x'}-1)}.
	\end{split}
	\end{equation}
	In the above, $n$ is the number of data point. $S_{-u}$ is the set of data points that returned by querying in $T$ using $-u$, and $|S_{-u}|$ denotes the number of elements in $S_{-u}$. $M_{x'}$ denotes the number of clusters in state $x'$. $C$ denotes the original component, $C_u$ and $C_v$ are the two new components after split with elements $u$ and $v$ in them. $K$ is the number of bits used for hashing, and $L$ is the number of hash tables probed. $Pr(-u,v)$ is the collision probability between $-u$ and $v$. We provide the derivations of the two formula in the supplementary material.
	\subsubsection{Naive LSH Smart-merge/Dumb-split}
	LSH based Merge begins by randomly selecting an element $u$ in the dataset. Then use LSS to sample from hash tables $T$ to get another element $v$ which is similar with $u$. Then, if the mixture component of $u$ and $v$ are different, then we do merge operation for the corresponding two mixture component.
	If $u$ and $v$ are in the same components, we do a dumb split: randomly select one cluster, and split this component into two separate components.
	
	We provide the the probability of the merge move $q(x'|x)$ and the corresponding inverse probability $q(x|x')$: 
	\begin{equation}
	\label{eq:lshmp}
	\begin{split}
	&q(x'|x) = \sum_{u}^{C_u} \sum_{v}^{C_v} \left( \frac{1}{n} \left(1-\left(1-Pr(u,v)^K\right)^L \right) \frac{|C_v \cap S_{u}|}{|S_{u}|} \right),\\&q(x|x') = \frac{1}{M_{x'}}(\frac{1}{2})^{|C_u| + |C_v|}.
	\end{split}
	\end{equation}
	
	In the above, $S_{u}$ is the set of data points that returned by query in $T$ using $u$. $|S_{u}|$ denotes the number of elements in $S^{s}$. All the other symbols have the same meaning as before. $Pr(u,v)$ is the collision probability between $u$ and $v$.
	The derivations of these formula is shown in the supplementary material.
	
	
	Noticing that, to calculate the transition probabilities in Eq. \ref{eq:lshsp} and Eq. \ref{eq:lshmp}, we need to sum over all possible $u$ and $v$ in the two components $C_u$ and $C_v$. This could be expensive when the cluster size is large. In other words, this complexity of this proposal is quadratic to the size of the cluster. It is still better than $O(n^2)$, but not quite there. It turns out, surprisingly,  that a very unique design of proposal eliminates this quadratic barrier. It is the unique mathematical properties of MinHash and a novel generalization of its $k$-way collision probability that makes this possible.
	\subsection{MinSM: MinHash based Split-Merge MCMC}
	\label{sec:minsm}
	
	LSH does similarity based sampling. Thus, we can sample pairs $u$ and $v$ in adaptive fashion efficiently. A split of cluster $C$ into $C_u$ and $C_v$ can happen because of any two elements $x \in C_u$ and $y \in C_v$ being samples. As a result, the transition probability requires accumulating non-uniform probabilities of all possible combinations, making it expensive to compute. It would be nice if we can design a split of where we can directly evaluate the probability expression without this costly accumulations.
	
	Ideally, after identifying $u$ we should split so that all the elements similar to $u$ goes to $C_u$ and rest goes to $C_v$. This will be a significantly more informative proposal than random assignments to $C_u$ and $C_v$. However, evaluating the transition probability of configuration under LSH would be impossible, or computationally intractable, as LSH sampling is correlated and the expressions are contrived. It turns out, that MinHash with a very specific design exactly achieves this otherwise impossible state with the cost of evaluating the transition probability linear in the size of the cluster. A unique property of MinHash is that we can compute, in closed form and linear cost, the probability of collision of a set of points of any size $\ge 2$. Such computation is not possible with any other know LSH including the popular random projections.
	
	We provide a novel extension of the collision probability of MinHash to also include the probability of collision with a given set and no collision with another given set (See Equation~\ref{eq:kwayCollProb}). It is surprising that despite many non-trivial correlations, the final probability expression is very simple and only requires linear computations. As a result, we can directly get the split of a cluster into two sets (or clusters) and at the same time compute the transition probability.  The novel design and analysis of Minhash, presented here, could be of independent interest in itself.
	
	
	
	
	
	\subsubsection{MinHash Smart-split/Dumb-merge}
	MinHash based split begins by flipping a coin to randomly choose from the action of Smart-split or Dumb merge.
	
	The LSH smart-split begins by randomly selecting an element $u$ in the dataset. Then, we use LSS (Locality-sensitive Sampler) to sample a set of points that are likely to be similar to $u$ from $T$, i.e., query $T$ with $u$. Here we use Weighted MinHash as the LSH and $K=1$ is necessary. $K \ge 1$ makes the probability computations out of reach. Instead of sampling a point from the bucket, as we do with LSS, we just report the whole bucket as the set. Let us denote this sampled set as $S_u$.  We now split the component $C_u$ into two components: $C_u \cap S_u, ~ C_u - S_u$. If the action is a dumb merge, then we randomly select two components and merge these two components into one component.
	
	Given a new state $x'$, and the corresponding old state $x$, we can precisely compute the probability of the proposed split move $q(x'|x)$ and the corresponding inverse move probability $q(x|x')$ as follow:
	
	
	Define $p$ is the probability of agreement of weighted minhash of $u$ with all of the data point in the queried set $S_u$. The known theory says that the expression of $p$ is given by Equation~\ref{eq:kwayCollProb}. However, we want something more, we want all elements of $S_u$ to collide with $u$ in the bucket and anything in $C_u -S_u$ to not collide. Define $Prob$ as the probability of agreement of weighted minhash of $u$ with all of $S_u$ and none of the data point in $C_u - S_u$. It turns our that we can calculate this probability exactly as:
	\begin{equation}
	\begin{split}
	Prob = \frac{\sum_{j}^{2D} \max\{0, (x^j_{\min} - x^j_{\max})\}}{\sum_{j}^{2D} x^j_{all}}.
	\end{split}
	\end{equation}
	
	where $x^j_{\min} = \min_{x \in C_u \cap S_u } \{x_j\}$, $x^j_{\max} = \max_{x \in C - S_u}\{x_j\}$ and $x^j_{all} =  \max_{x \in C_u}\{x_j\}$
	
	When we only use $K=1$ Minhash, then the corresponding proposal distribution is shown as follow:
	\begin{equation}
	\begin{split}
	q(x'|x) = \frac{|S_u|}{n}\times Prob.
	\end{split}
	\end{equation}
	
	\begin{figure}[t]
		\centering
		\includegraphics[width=0.5\textwidth]{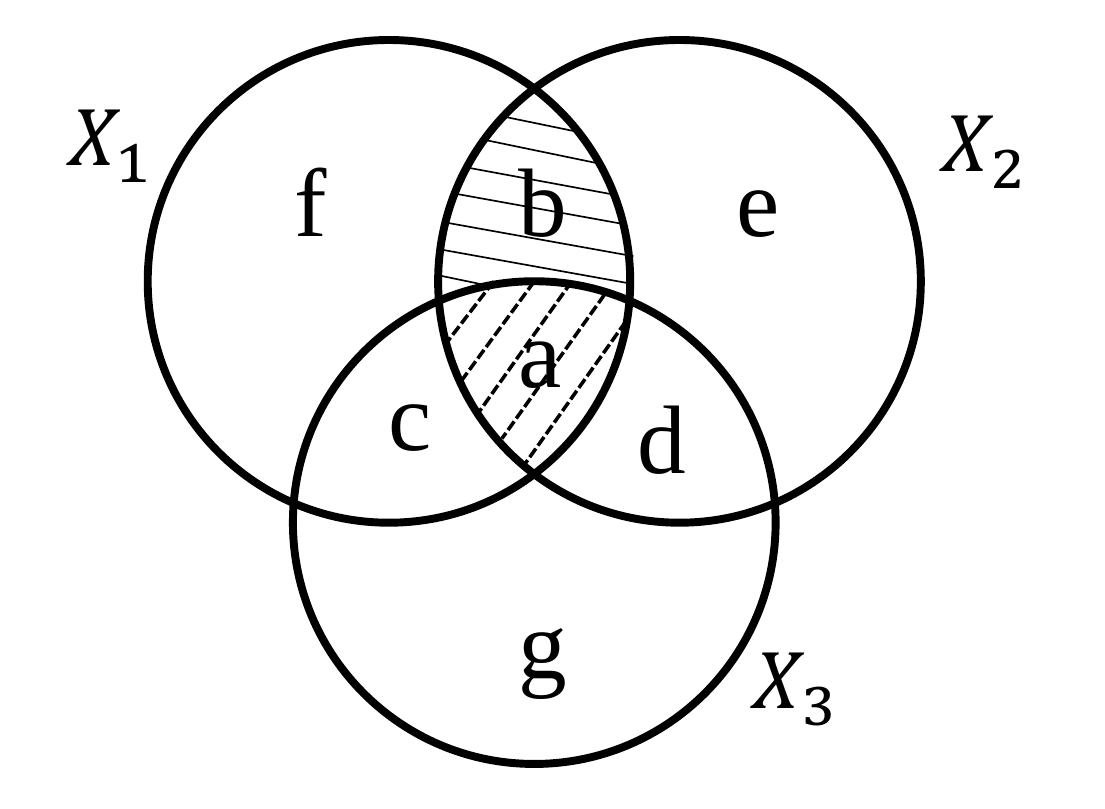}
		\caption{Three way Minwise Hashing.}
		\label{fig:threeway}
	\end{figure}
	
	For the sake of understanding, we give an illustration of the proof. Consider Figure~\ref{fig:threeway}. Let's start with vanilla MinHash over sets and the arguments will naturally extend to weighted versions.   
	Given $X_1$, $X_2$ and $X_3$. We want the probability that the MinHash of $X_1$ and $X_2$ collide but not of $X_3$. From the theory of consistent sampling~\cite{shrivastava2013beyond,shrivastava2016simple,manasse2010consistent}. This will happen if we sample from $b$ and the possibility is the union.  Thus the probability is $\frac{b}{a+b+c+d+e+f+g} = \frac{|X_1 \cap X_2| - |X_3|}{|X_1 \cup X_2 \cup X_3|}$ which is essentially we want the minimum of $|X_1 \cup X_2 \cup X_3|$ to be  sampled from the intersection of $X_1$ and $X_2$ and not from $X_3$. That is the only way the MinHash of $X_1$ and $X_2$ will agree but not of $X_3$. This argument can be naturally extendend if we want $X_1, X_2, ..., X_h$ to have same minhash and not $Y_1, Y_2, ...,Y_g$, the probability can be written as:
	
	$$\frac{\max\{0, |X_1 \cap X_2 \cap ... \cap X_h| - |Y_1 \cup Y_2 \cup ... \cup Y_g|\}}{| X_1 \cup X_2 \cup ... \cup X_h \cup Y_1 \cup Y_2 \cup ... \cup Y_g|}. $$
	
	Now for weighted sets (non-binary), we can replace intersection with minimum and unions with max leading to the desired expression, which is due to the seminal works in consistent weighted sampling a strict generalization of MinHash. See~\cite{shrivastava2013beyond,shrivastava2016simple,manasse2010consistent} for details.  Also using~\cite{leskovec2014mining} we can extend it to negative weights as well using simple feature transformation.
	
	It should be noted that this expression only requires cost linear in the size of the cluster $C_u$ being split.
	With this value of $Prob$, the corresponding transition probability for the split move is:
	\begin{equation}
	\label{eq:lshspn}
	\begin{split}
	q(x'|x) = \frac{|S_u|}{n}\times Prob, ~~~ q(x|x') = \frac{2}{M_{x'}(M_{x'}-1)}.
	\end{split}
	\end{equation}
	
	In the above, $n$ is the number of data point. $S_{u}$ is the set of data points that returned by querying in $T$ using $u$, and $|S_{u}|$ denotes the number of elements in $S_{u}$.
	$D$ is the dimension of the data.
	$M_{x'}$ denotes the number of clusters in state $x'$.
	
	To be able to compute this expression and also get an informative split was the primary reason for many choices that we made. For example, $K=1$ us needed so that we can compute $Prob$ in a simple closed form. As a result, we obtain a very unique proposal. The idea and design could be of independent interest in itself.
	Please refer to supplementary for the derivation of these equations.
	
	\subsubsection{Minhash Smart-merge/Dumb-split}
	The proposed smart-merge begins by randomly selecting a center $u$ in the dataset. Then, we use LSS (Locality-sensitive Sampler) to sample a center $v$ that are likely to be similar to $u$. Then we merge the component $C_{u}$ and $C_{v}$ to one component.
	
	If the action is a dumb split: randomly select one cluster, and split this component into two separate components uniformly.
	
	Given a new state $x'$, and the corresponding old state $x$. We provide the probability of the merge move $q(x'|x)$ and the corresponding inverse probability $q(x|x')$ as follow:
	
	\begin{equation}
	\begin{split}
	\label{eq:lshmpn}
	&q(x'|x) = \frac{1}{M_{x}}\frac{\sum_{j}^{2D} \min\{{u}_j, {v}_j\}}{\sum_{j}^{2D} \max\{{u}_j, {v}_j\}} \frac{1}{|S_{u}|}, \\& q(x|x') = \frac{1}{M_{x'}}(\frac{1}{2})^{|C_u| + |C_v|}.
	\end{split}
	\end{equation}
	
	
	In the above, $S_{u}$ is the set of data points that returned by the query in hash table $T$ using $u$. $|S_{u}|$ denotes the number of elements in $S_{u}$. $u_j$ denotes $j$-th feature of the data point $u$. All the other symbols have the same meaning as before.

	As we introduced before, our proposed algorithm belongs to the general framework of metroplis-hastings algorithm \cite{andrieu2003introduction}. After each split/merge move, we need to calculate the acceptance rate $\alpha (x'|x)$ for this move which is given by: $
	\alpha (x'|x) = \min \{1, \frac{L(x')q(x|x')}{L(x)q(x'|x)}\}
	$, where $x'$ is the proposed new state, $x$ is the previous state, $q(x'|x)$ here is the designed proposal distribution, and it can be calculated as introduced in previous sections. $\mathcal{L}(x)$ is the likelihood value of the state $x$.
	
	The likelihood of the data is generally in the form of $L(x) = \prod_{D} p_{j}(e_i)$, where $p_{j}(e_i)$ is the probability of $e_i \in D$ in it's corresponding component $C_j$. $D$ denotes the total dataset. In the split merge MCMC, only the components that being split/merged will change of the likelihood value. 
	So, that the ratio $\frac{L(x')}{L(x)}$ is cheap to compute, since all the probability of unchanged data will be canceled.

\section{Empirical Study}
\subsection{Experiment Setup}
		\begin{figure}[t]
		\centering
		\begin{subfigure}[b]{0.39\textwidth}
			\includegraphics[width=\textwidth]{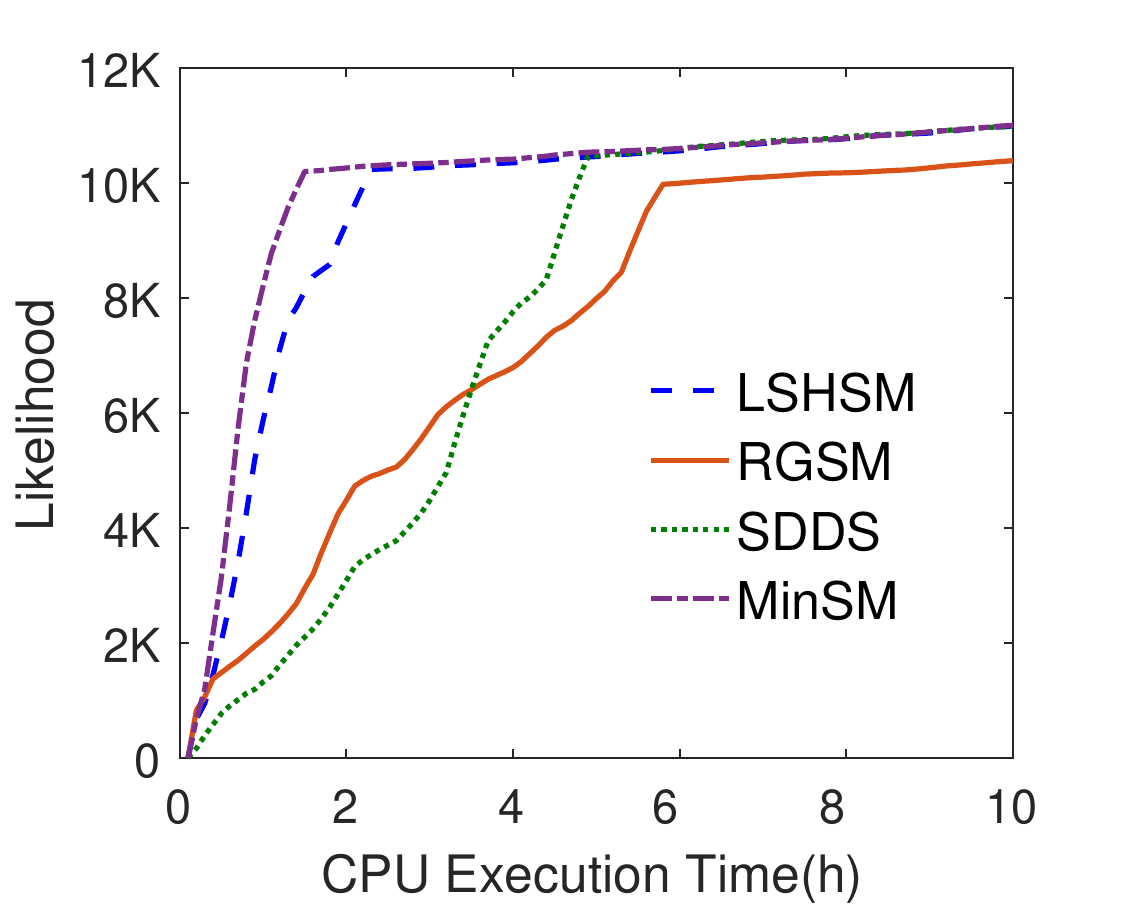}
			\caption{KDDCUP Dataset}
			\label{fig:timekdd}
		\end{subfigure}
		\begin{subfigure}[b]{0.39\textwidth}
			\includegraphics[width=\textwidth]{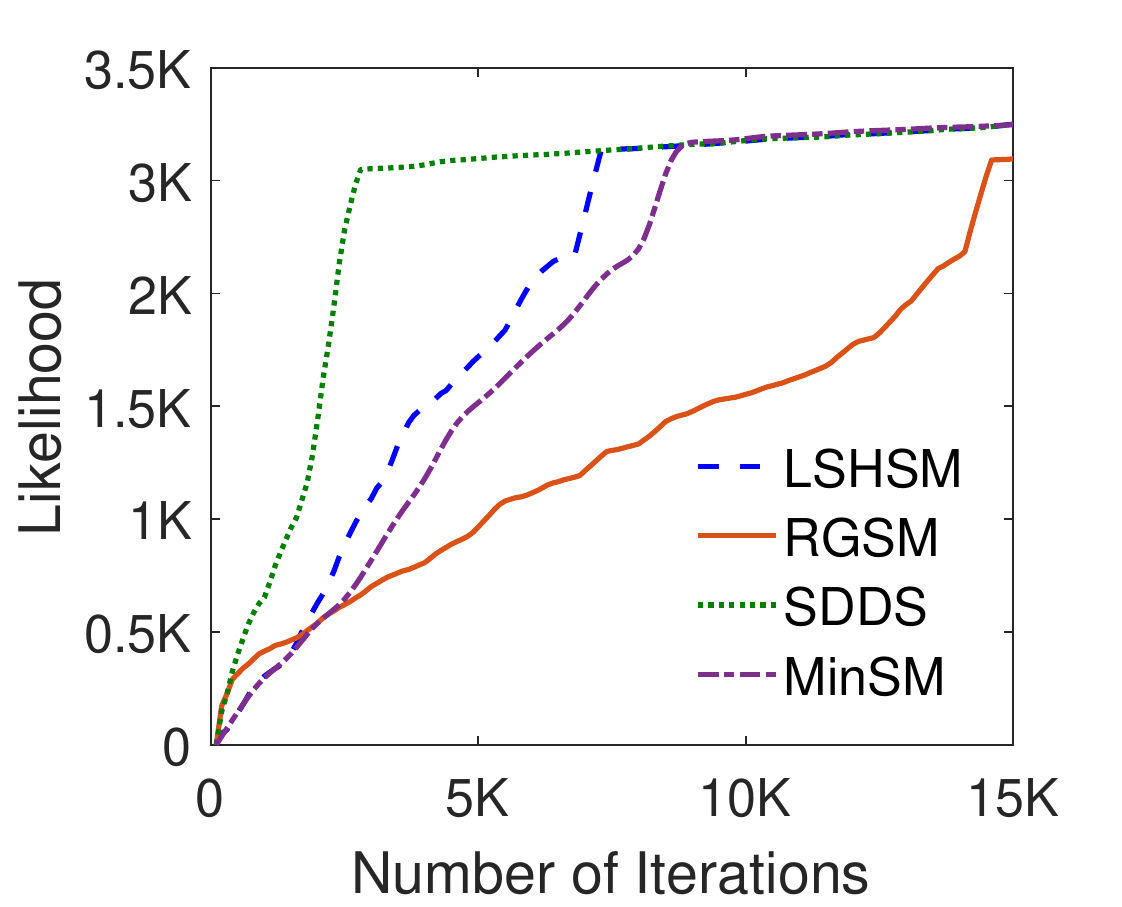}
			\caption{KDDCUP Dataset}
			\label{fig:epochkdd}
		\end{subfigure} 
		~ 
		\begin{subfigure}[b]{0.39\textwidth}
			\includegraphics[width=\textwidth]{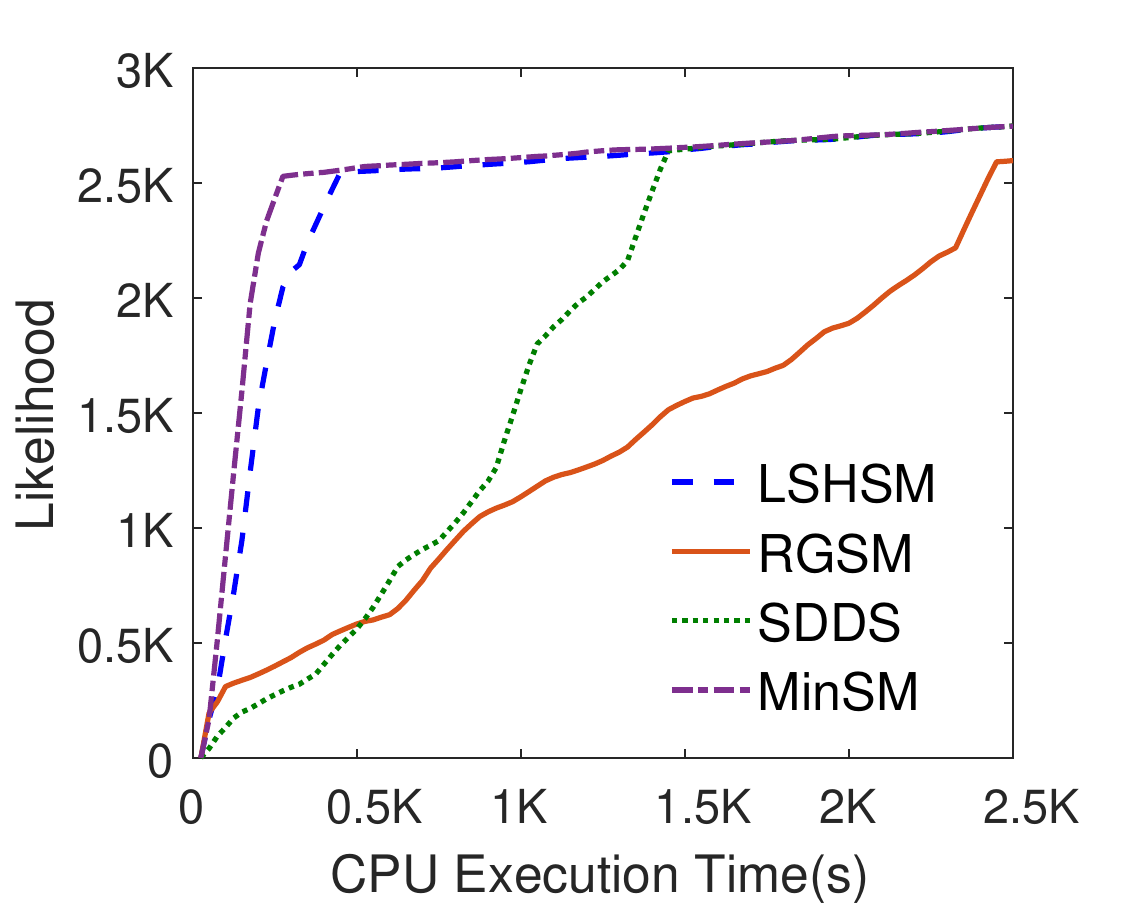}
			\caption{PubMed Dataset}
			\label{fig:timepub}
		\end{subfigure}
		~ 
		\begin{subfigure}[b]{0.39\textwidth}
			\includegraphics[width=\textwidth]{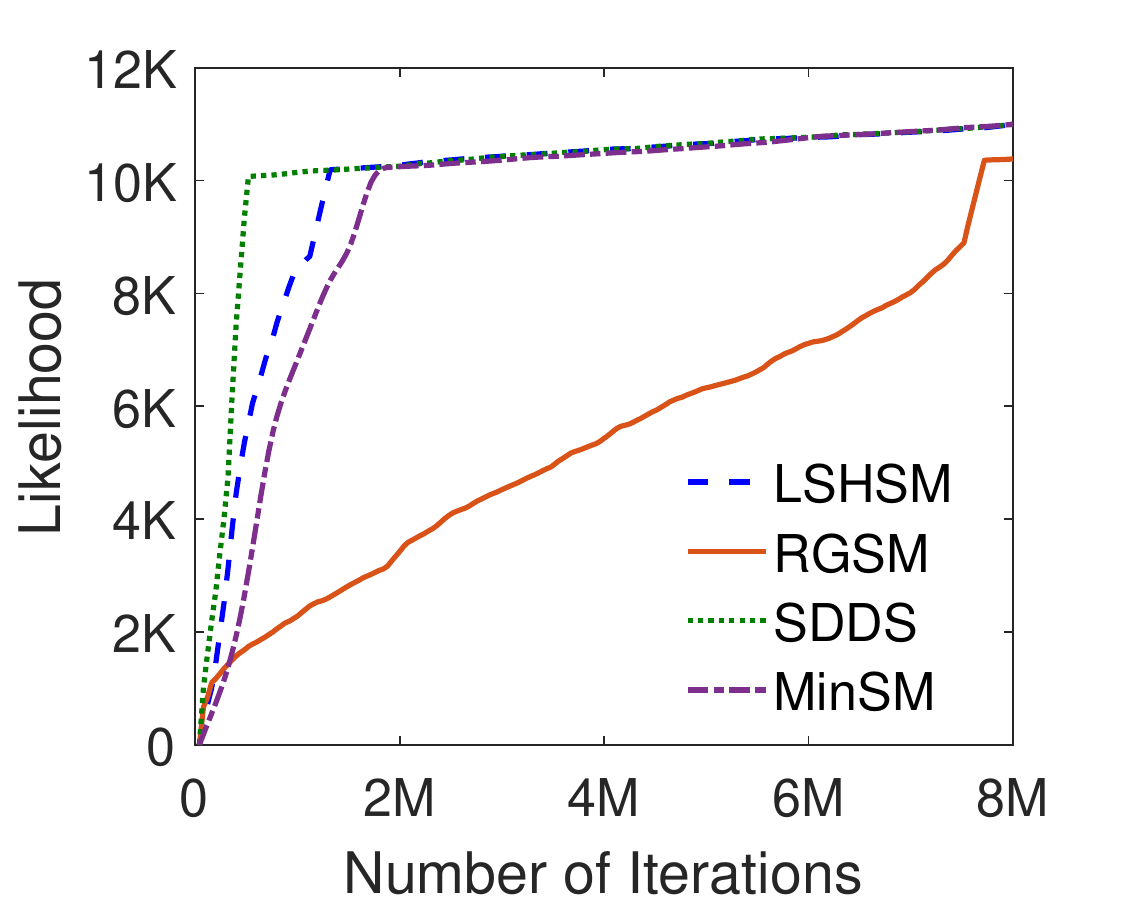}
			\caption{PubMed Dataset}
			\label{fig:epochpub}
		\end{subfigure}
		\caption{The time and iteration wise comparison of the likelihood for difference methods on the two real dataset. It is obviously that our proposed MinSM algorithm can be at least $6$ times faster than the state of the art algorithms in the real large dataset.} 
		\label{fig:realtime}
	\end{figure}
	\label{sec:exp}
	In this section, we demonstrate the advantage of our proposed models by applying it to the Gaussian Mixture model inference and compare it with state-of-the-art sampling methods.
	\subsection{Gaussian Mixture Model}
	We briefly review the Gaussian Mixture Model. A Gaussian mixture density is a weighted sum of component densities.
	For a $M$-class clustering task, we could have a set of GMMs associated with each cluster. For a $D$-dimensional feature vector denoted as $x$, the mixture density is defined as $p(x) = \sum_{i=1}^{M} w_ip_i(x)$, where $w_i, i=1,...,M$ are the mixture weights which satisfy the constraint that $\sum_{i}^{M} = 1$ and $w_i \ge 0$. The mixture density is a weighted linear combination of component $M$ uni-model Gaussian density functions $p_i(x), i=1,...,M$.  The Gaussian mixture density is parameterized by the mixture weights, mean vectors and covariance vectors from all components densities.
	
	For a GMM-based clustering task, the goal of the model training is to estimate the parameters of the GMM so that the Gaussian mixture density can best match the distribution of the training feature vectors. Estimating the parameters of the GMM  using the expectation-maximization (EM) algorithm~\cite{nasrabadi2007pattern} is popular. However, in most of the real world applications, the number of clusters $M$ is not known, which is required by the EM algorithm. On the other hand, Split-Merge based MCMC algorithms are used for inference when $M$ is unknown, which is also the focus of this paper. We therefore only compare our proposal LSHSM and other state-of-the-art split-merge algorithms on GMM clustering which does not require the prior knowledge of the number of clusters.
	
	\textbf{Competing Algorithms:} We compare following four split-merge MCMC sampling algorithm on GMM with an unknown number of clusters:
	\textbf{RGSM:} Restricted Gibbs split-merge MCMC algorithm \cite{jain2004split} is considered as one of the state-of-the-art sampling algorithm.
	\textbf{SDDS:} Smart-Dumb/Dumb-Smart Split Merge algorithm \cite{wang2015smart}.
	\textbf{LSHSM:} The Naive version of LSH based Split Merge algorithm by using Sign Random Projection. In the LSHSM method, we use fixed $K=10$ and $L=10$ for all the dataset. We fix the hashing scheme to be signed random projection.
	\textbf{MinSM:} LSH based split merge algorithm is the proposed method in this paper. In the MinSM method, we use fixed $K=1$ and $L=1$ for all the dataset.
	
		\begin{figure}[!htbp]
		\centering
		\begin{subfigure}[b]{0.39\textwidth}
			\includegraphics[width=\textwidth]{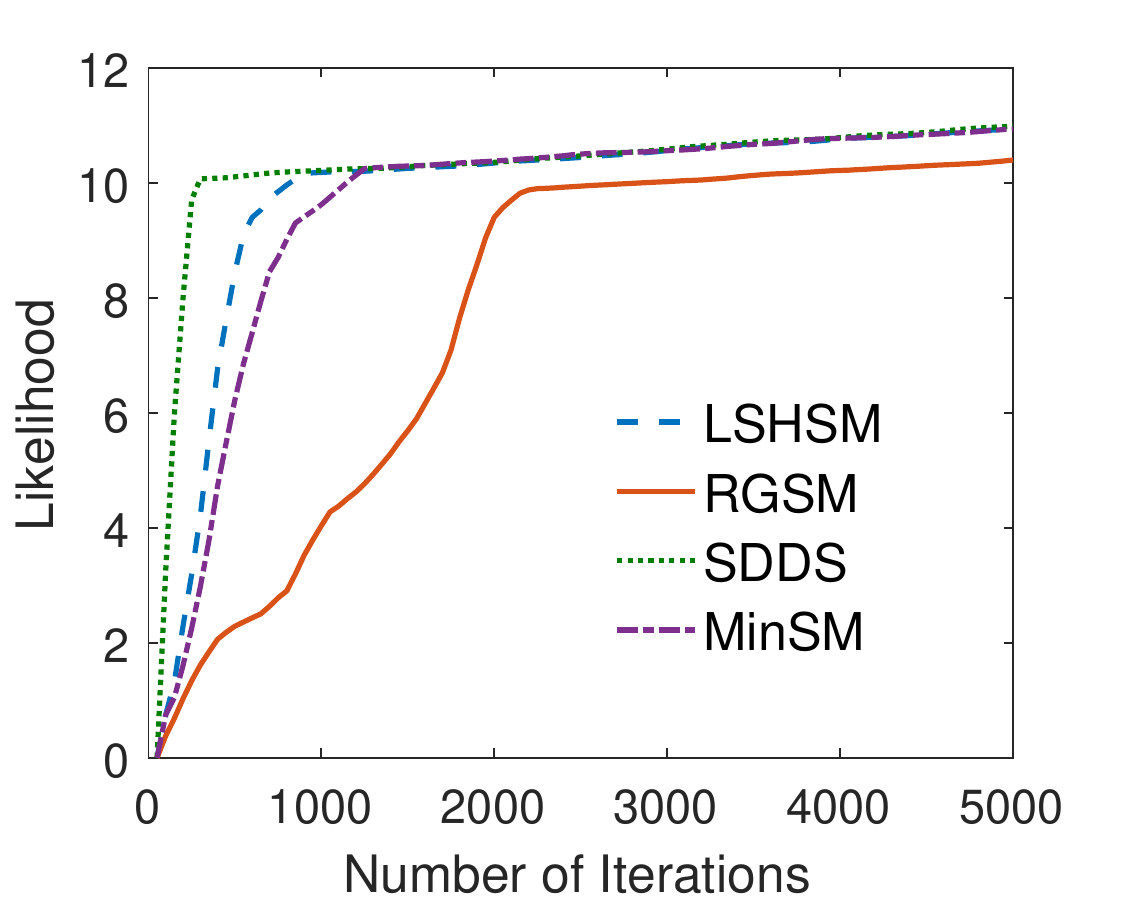}
			\caption{S1 dataset}
			\label{fig:epoch100}
		\end{subfigure}
		~ 
		\begin{subfigure}[b]{0.39\textwidth}
			\includegraphics[width=\textwidth]{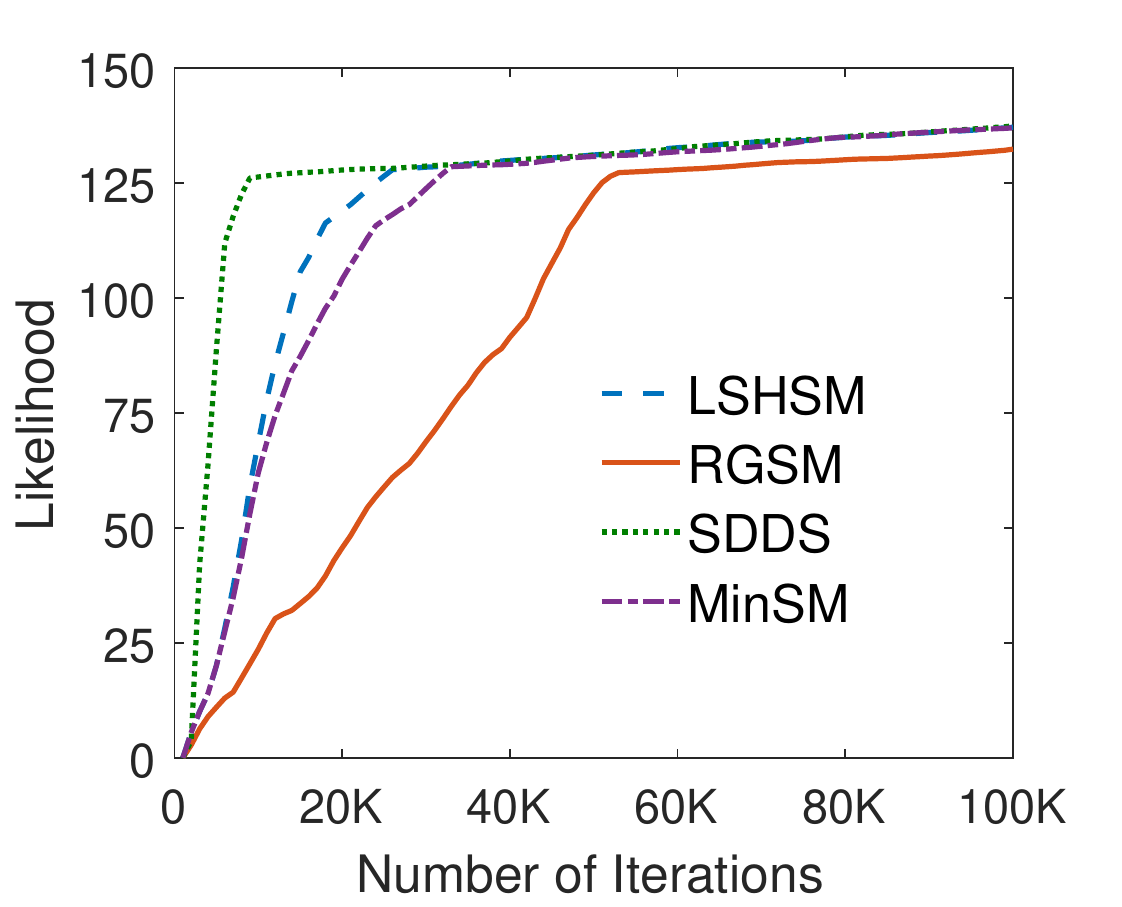}
			\caption{S2 dataset}
			\label{fig:epoch1000}
		\end{subfigure}
		~ 
		\begin{subfigure}[b]{0.39\textwidth}
			\includegraphics[width=\textwidth]{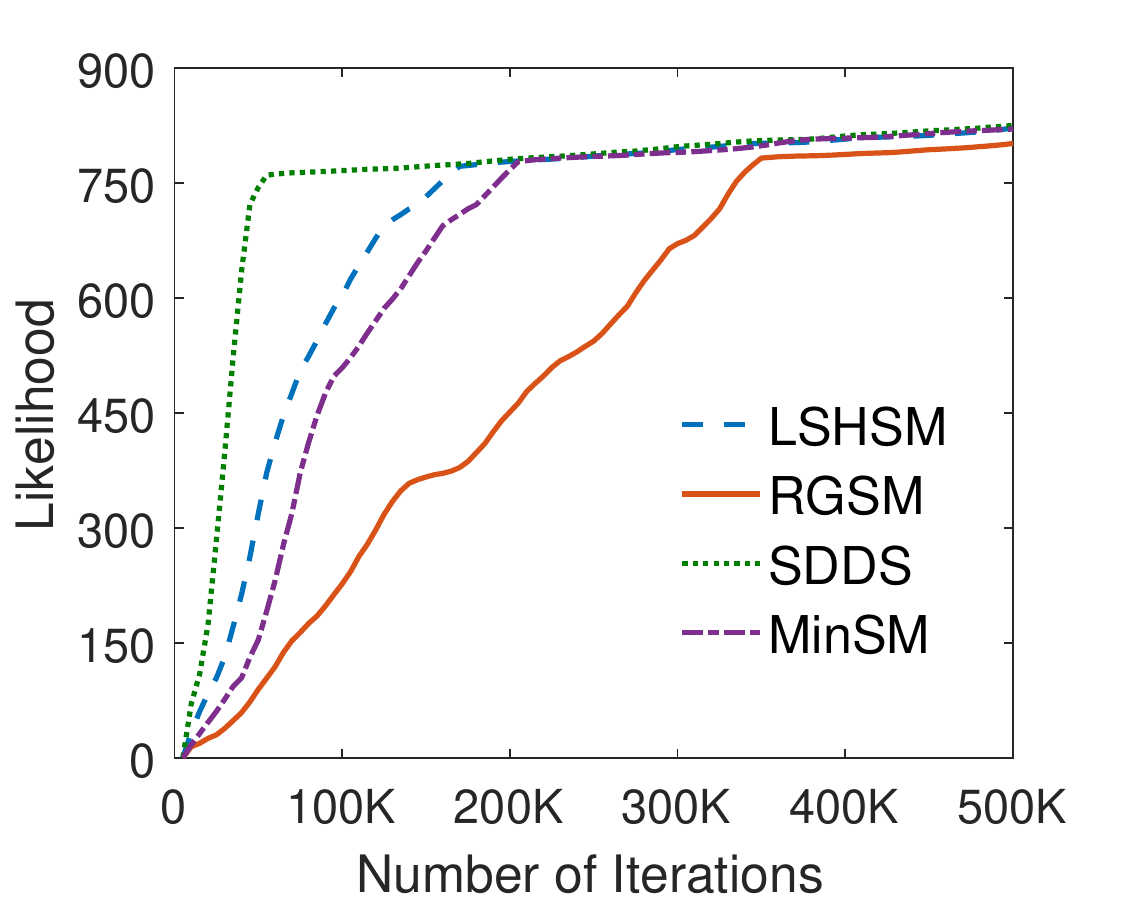}
			\caption{S3 Dataset}
			\label{fig:epoch10000}
		\end{subfigure}
		\begin{subfigure}[b]{0.39\textwidth}
			\includegraphics[width=\textwidth]{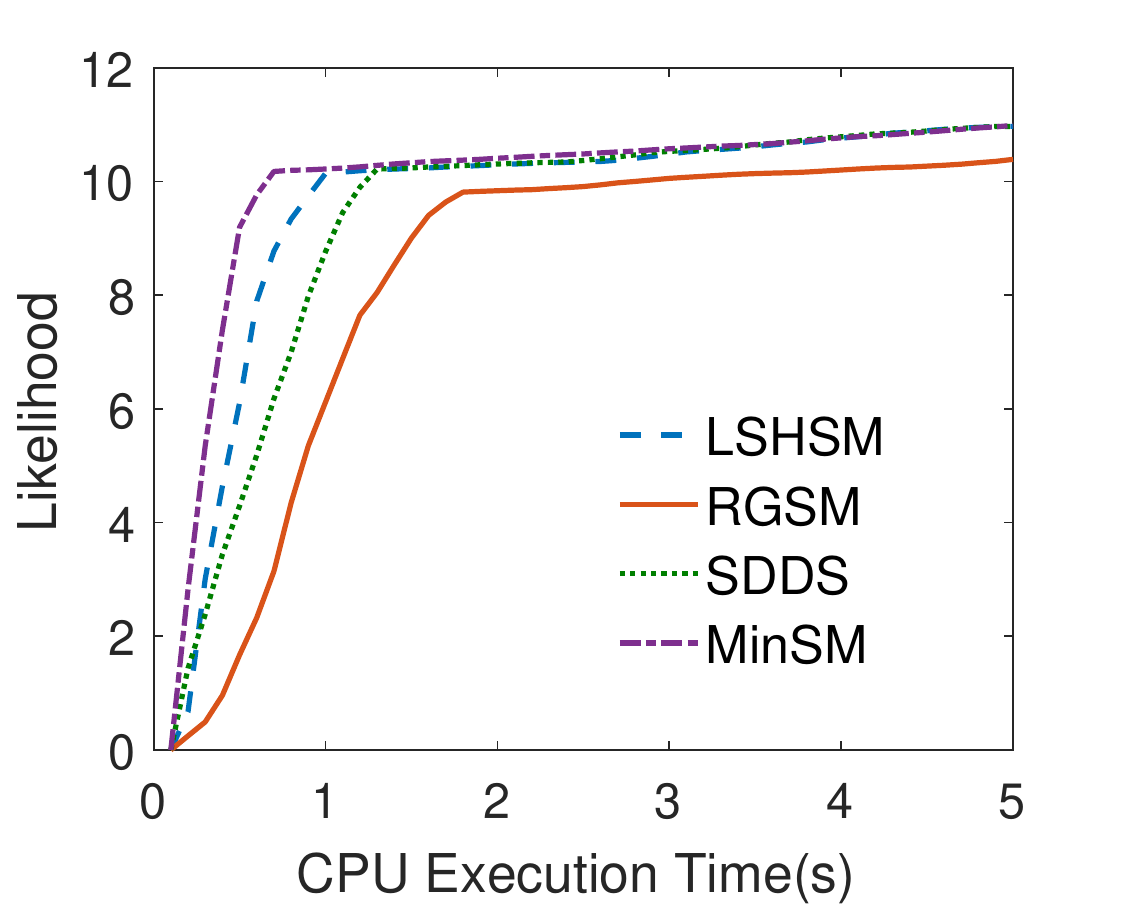}
			\caption{S1 dataset}
			\label{fig:time100}
		\end{subfigure}
		~ 
		\begin{subfigure}[b]{0.39\textwidth}
			\includegraphics[width=\textwidth]{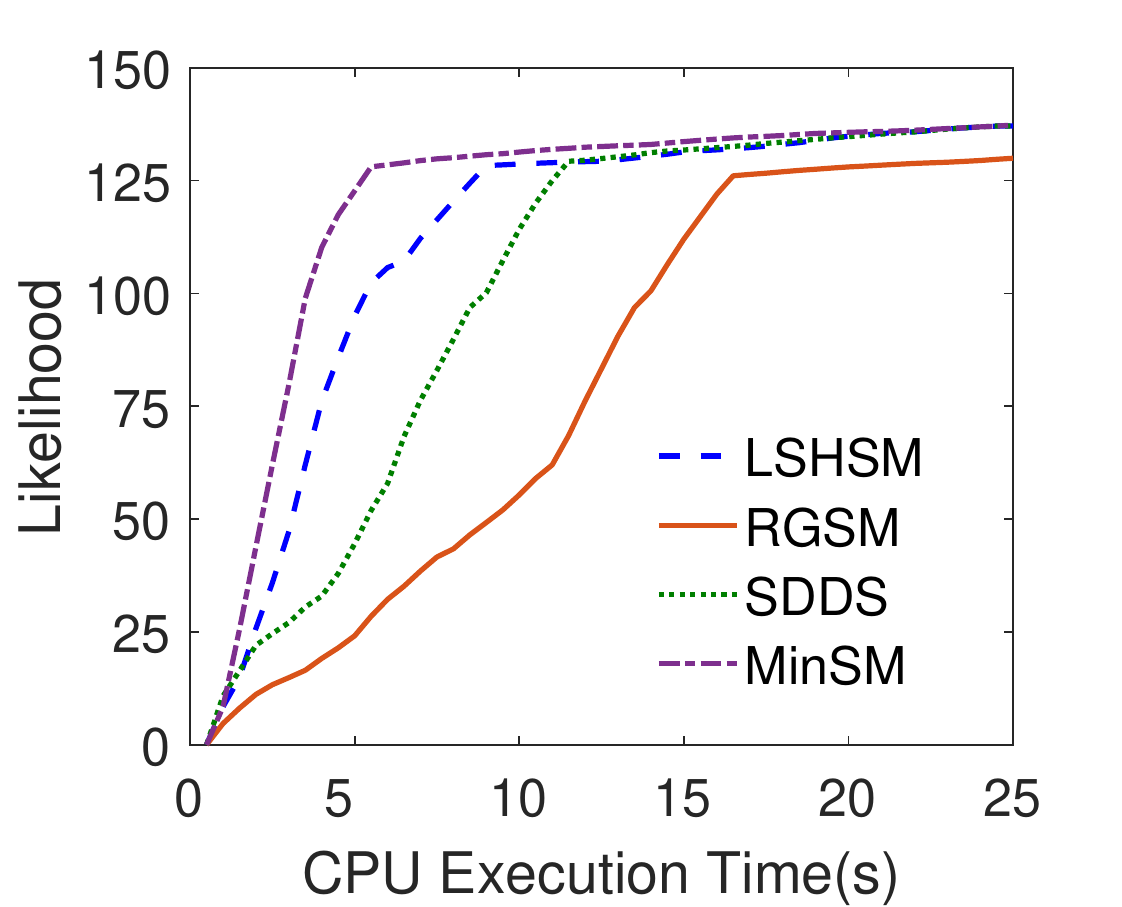}
			\caption{S2 dataset}
			\label{fig:time1000}
		\end{subfigure}
		~ 
		\begin{subfigure}[b]{0.39\textwidth}
			\includegraphics[width=\textwidth]{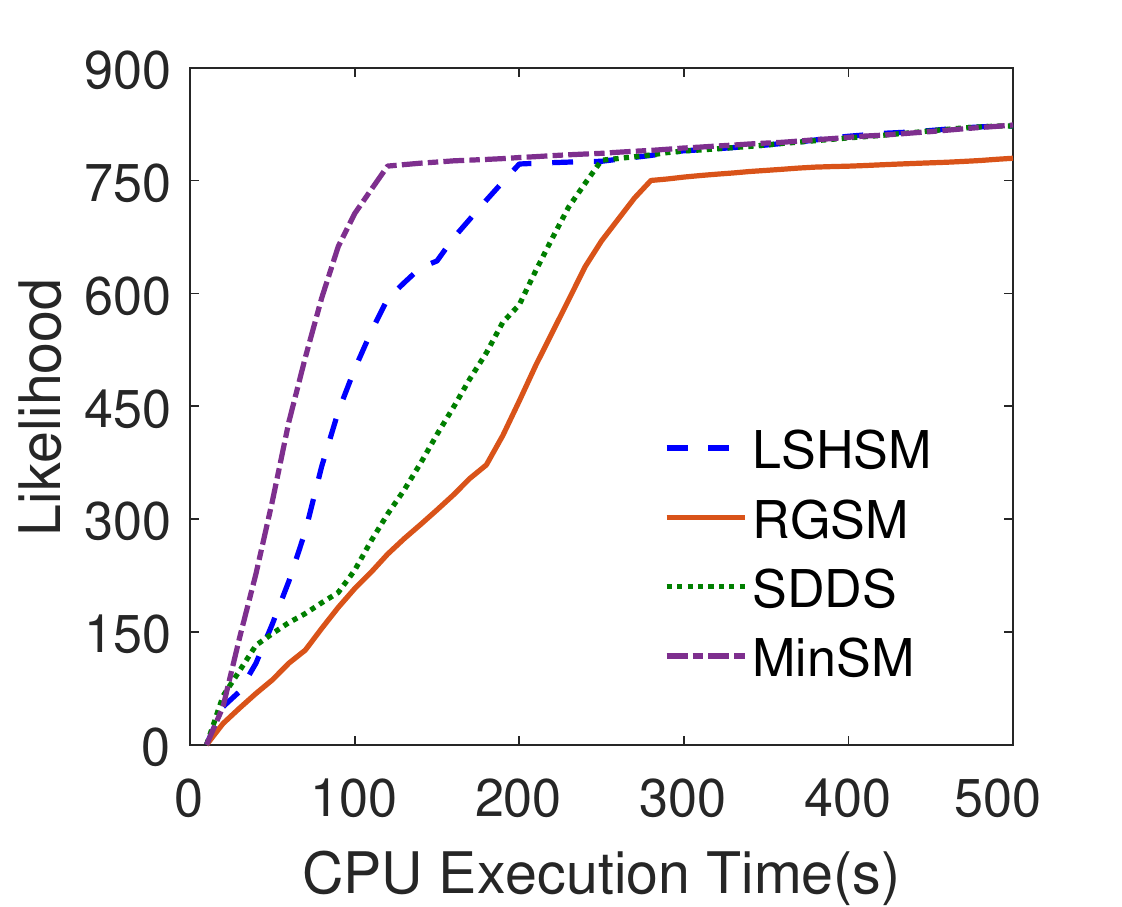}
			\caption{S3 dataset}
			\label{fig:time10000}
		\end{subfigure}
		\caption{The time and iteration wise comparison of the likelihood for difference methods on the Synthetic Dataset. MinSM outperforms the other baselines by a large margin. It is also clear that requiring less iteration does not mean faster convergence. }
		\label{fig:syntime}
	\end{figure}
	
	\textbf{Dataset:} We evaluate the effectiveness of our algorithm on both two large real-world datasets: \textbf{KDDCUP} and \textbf{PubMed}.
	\textbf{KDDCUP} data was used in the KDD Cup 2004 data mining competition. It contains $145751$ data point. The dimensionality of the dataset is $74$. We have $2000$ ground truth cluster labels for this dataset.   \footnote{https://cs.joensuu.fi/sipu/datasets/}
	The \textbf{PubMed} abstraction dataset contains $8200000$ abstractions that extracted from the PubMed \footnote{www.pubmed.gov}. All the documents represented as the bag-of-words representation. In the data set, we have $141043$, different words. This data set is ideal for document clustering or topic modeling. 
	
	Synthetic data is a standard way of testing GMM models \cite{nasrabadi2007pattern}. So, in this paper, we also use synthetic datasets as a sanity check to evaluate the performance of different methods.
	The process of generating the synthetic dataset is as follow:
	Randomly generate $k$ different Gaussian distributions (with different corresponding mean and variance). We fix the $k=10$ in our experiment.
	Then based on the randomly generated Gaussian distributions, we generate a set of data points for each Gaussian distribution. Here we fix the dimensionality of each data point to $25$.
	In this experiment, we generate three sythntic dataset with different size (e.g. $100$, $1000$, $10000$). We name the three synthetic dataset as \textbf{S1}, \textbf{S2}, \textbf{S3}.
	
	

	
	\subsection{Speed Comparison and Analysis}
	We first plot the evolution of likelihood both as a function of iterations as well as the time of all the three competing methods. The evolution of likelihood and time with iterations on two real-world data is shown in Fig. \ref{fig:realtime}. The result on three synthetic data set is shown in Fig. \ref{fig:syntime}.
	
	We can see a consistent trend in the evolution of likelihood, which holds true for both simulated as well as real datasets.  First of all, RGSM consistently performs poorly and requires both more iterations as well as time. The need of combining smart and dumb moves for faster convergence made in~\cite{wang2015smart}, seems necessary. RGSM does not use it and hence leads to poor, even iteration wise, convergence.
	
	
	SDDS seems to do quite well, compared to our proposed LSHSM when we look at iteration wise convergence. However, when we look at the time, the picture is completely changed. MinSM is significantly faster than SDDS, even if the convergence is slower iteration wise. This is not surprising because the per-iteration cost of MinSM  is orders of magnitude less than SDDS. SDDS hides the computations inside the iteration by evaluating every possible state in each iteration, based on likelihood,  is equivalent to several random iterations combined.  Such costly evaluation per iteration can give a false impressing of less iteration.
	
	It is clear from the plots that merely comparing iterations and acceptance ratio can give a false impression of superiority.  Time wise comparison is a legitimate comparison of overall computational efficiency. Clearly, MinSM outperforms the other baselines by a large margin.
	
	
	\subsection{Clustering Accuracy Comparison}
	
	\begin{table}
		\centering
		\caption{Clustering Accuracy for Different Methods}
		\label{tab:clusacc}
		\begin{tabular}{ccccccc}
			\hline
			Methods & Metric & S1 & S2 & S3 & KDD & Pub \\
			\hline
			RGSM &NMI  & 0.96 & 0.93 & 0.88 & 0.74 & 0.63\\
			~ & Accuracy & 0.95 & 0.92 & 0.87 & 0.68 & 0.62 \\
			\hline
			SDDS & NMI & \textbf{0.97} & \textbf{0.96} & 0.95 & \textbf{0.86} & \textbf{0.80}\\
			~ & Accuracy & \textbf{0.98} & \textbf{0.97} & 0.94 & \textbf{0.85} & \textbf{0.77} \\
			\hline
			LSHSM & NMI & 0.96 & 0.95 & \textbf{0.96} & 0.84 & 0.77 \\
			~ & Accuracy& 0.97 & 0.94 & 0.96 & 0.83 & 0.75 \\
			\hline		
			MinSM & NMI & 0.96 & 0.94 & \textbf{0.96} & 0.83 & 0.75 \\
			~ & Accuracy& 0.97 & 0.94 & \textbf{0.97} & 0.84 & 0.74 \\
			\hline		
		\end{tabular}
	\end{table}
	
	To evaluate the clustering performance of different algorithms, we use two widely used measures (Accuracy and NMI \cite{nasrabadi2007pattern}). 
	\textbf{Normalized Mutual Information (NMI)} \cite{nasrabadi2007pattern} is widely used for measuring the performance of clustering algorithms. It can be calculated as $NMI (C,C') = \frac{I(C;C')}{\sqrt{H(C)H(C')}},$
	where $H(C)$ and $H(C')$ are the marginal entropies, $I(C;C')$ is the mutual information between $C'$ and $C$.
	The \textbf{Accuracy} measure, which is calculated as the percentage of target objects going to the correct cluster, is defined as $Accuracy = \frac{\sum_{i=1}^{k} a_i}{n},$ where $a_i$ is the number of data objects clustered to its corresponding true cluster, $k$ is the number of cluster and $n$ is the number of data objects in the dataset.
	
	Table \ref{tab:clusacc} shows the clustering accuracy of different competing methods.
	We can see that the MinSM, LSHSM and SDDS are much more accurate than RGSM. This observation is in agreement with the likelihood plots.
	On the other hand, the accuracy difference between MinSM, LSHSM and SDDS is negligible.

	\section{Conclusion}
	The Split-Merge MCMC (Monte Carlo Markov Chain) is one of the essential and popular variants of MCMC for problems with an unknown number of components. It is a well known that the inference process of SplitMerge
	MCMC is computational expensive which is not applicable for the large-scale dataset. Existing approaches that try to speed up the split-merge MCMC are stuck in a computational chicken-and-egg loop problem.
	
	In this paper, we proposed MinSM, accelerating Split Merge MCMC via weighted Minhash. The new splitmerge MCMC has constant time update, and at the same time the proposal is informative and needs significantly fewer iterations than random split-merge. Overall, we obtain a sweet tradeoff between convergence and per update cost. Experiments with Gaussian Mixture Model on two real-world datasets demonstrate much faster convergence and better scaling to large datasets.
	
	\bibliography{mcmc}

\begin{thebibliography}{}

\bibitem[\protect\citename{Andrieu {\em et~al.\ }\relax,
  }2003]{andrieu2003introduction}
Andrieu, Christophe, De~Freitas, Nando, Doucet, Arnaud, \& Jordan, Michael~I.
  2003.
\newblock An introduction to MCMC for machine learning.
\newblock {\em Machine learning}, {\bf 50}(1-2), 5--43.

\bibitem[\protect\citename{Bengio {\em et~al.\ }\relax, }2010]{bengio2010label}
Bengio, Samy, Weston, Jason, \& Grangier, David. 2010.
\newblock Label embedding trees for large multi-class tasks.
\newblock {\em Pages  163--171 of:} {\em Advances in Neural Information
  Processing Systems}.

\bibitem[\protect\citename{Chang \& Fisher~III, }2013]{chang2013parallel}
Chang, Jason, \& Fisher~III, John~W. 2013.
\newblock Parallel sampling of DP mixture models using sub-cluster splits.
\newblock {\em Pages  620--628 of:} {\em Advances in Neural Information
  Processing Systems}.

\bibitem[\protect\citename{Charikar \& Siminelakis, }2017]{charikarhashing}
Charikar, Moses, \& Siminelakis, Paris. 2017.
\newblock Hashing-based-estimators for kernel density in high dimensions.
\newblock FOCS.

\bibitem[\protect\citename{Eronen {\em et~al.\ }\relax,
  }2003]{eronen2003markov}
Eronen, Lauri, Geerts, Floris, \& Toivonen, Hannu. 2003.
\newblock A Markov chain approach to reconstruction of long haplotypes.
\newblock {\em Pages  104--115 of:} {\em Biocomputing 2004}.
\newblock World Scientific.

\bibitem[\protect\citename{Gionis {\em et~al.\ }\relax,
  }1999]{gionis1999similarity}
Gionis, Aristides, Indyk, Piotr, Motwani, Rajeev, {\em et~al.\ }\relax. 1999.
\newblock Similarity search in high dimensions via hashing.
\newblock {\em Pages  518--529 of:} {\em VLDB},  vol. 99.

\bibitem[\protect\citename{Huelsenbeck \& Ronquist,
  }2001]{huelsenbeck2001mrbayes}
Huelsenbeck, John~P, \& Ronquist, Fredrik. 2001.
\newblock MRBAYES: Bayesian inference of phylogenetic trees.
\newblock {\em Bioinformatics}, {\bf 17}(8), 754--755.

\bibitem[\protect\citename{Hughes {\em et~al.\ }\relax,
  }2012]{hughes2012effective}
Hughes, Michael~C, Fox, Emily, \& Sudderth, Erik~B. 2012.
\newblock Effective split-merge monte carlo methods for nonparametric models of
  sequential data.
\newblock {\em Pages  1295--1303 of:} {\em Advances in neural information
  processing systems}.

\bibitem[\protect\citename{Jain \& Neal, }2004]{jain2004split}
Jain, Sonia, \& Neal, Radford~M. 2004.
\newblock A split-merge Markov chain Monte Carlo procedure for the Dirichlet
  process mixture model.
\newblock {\em Journal of Computational and Graphical Statistics}, {\bf 13}(1),
  158--182.

\bibitem[\protect\citename{Leskovec {\em et~al.\ }\relax,
  }2014]{leskovec2014mining}
Leskovec, Jure, Rajaraman, Anand, \& Ullman, Jeffrey~David. 2014.
\newblock {\em Mining of massive datasets}.
\newblock Cambridge university press.

\bibitem[\protect\citename{Li, }2017]{li2017linearized}
Li, Ping. 2017.
\newblock Linearized GMM kernels and normalized random fourier features.
\newblock {\em Pages  315--324 of:} {\em Proceedings of the 23rd ACM SIGKDD
  International Conference on Knowledge Discovery and Data Mining}.
\newblock ACM.

\bibitem[\protect\citename{Luo \& Shrivastava, }2017]{luo2017arrays}
Luo, Chen, \& Shrivastava, Anshumali. 2017.
\newblock Arrays of (locality-sensitive) Count Estimators (ACE): High-Speed
  Anomaly Detection via Cache Lookups.
\newblock {\em arXiv preprint arXiv:1706.06664}.

\bibitem[\protect\citename{Manasse {\em et~al.\ }\relax,
  }2010]{manasse2010consistent}
Manasse, Mark, McSherry, Frank, \& Talwar, Kunal. 2010.
\newblock Consistent weighted sampling.
\newblock {\em Unpublished technical report) http://research. microsoft.
  com/en-us/people/manasse}, {\bf 2}.

\bibitem[\protect\citename{Medvedovic {\em et~al.\ }\relax,
  }2004]{medvedovic2004bayesian}
Medvedovic, Mario, Yeung, Ka~Yee, \& Bumgarner, Roger~Eugene. 2004.
\newblock Bayesian mixture model based clustering of replicated microarray
  data.
\newblock {\em Bioinformatics}, {\bf 20}(8), 1222--1232.

\bibitem[\protect\citename{Nasrabadi, }2007]{nasrabadi2007pattern}
Nasrabadi, Nasser~M. 2007.
\newblock Pattern recognition and machine learning.
\newblock {\em Journal of electronic imaging}, {\bf 16}(4), 049901.

\bibitem[\protect\citename{Sharma \& Adlakha, }2015]{sharma2015computational}
Sharma, Amit, \& Adlakha, Neeru. 2015.
\newblock A Computational Model to Study the Concentrations of DNA, mRNA and
  Proteins in a Growing Cell.
\newblock {\em Journal of Medical Imaging and Health Informatics}, {\bf 5}(5),
  945--950.

\bibitem[\protect\citename{Shrivastava, }2016]{shrivastava2016simple}
Shrivastava, Anshumali. 2016.
\newblock Simple and efficient weighted minwise hashing.
\newblock {\em Pages  1498--1506 of:} {\em Advances in Neural Information
  Processing Systems}.

\bibitem[\protect\citename{Shrivastava \& Li, }2013]{shrivastava2013beyond}
Shrivastava, Anshumali, \& Li, Ping. 2013.
\newblock Beyond Pairwise: Provably Fast Algorithms for Approximate $ k $-Way
  Similarity Search.
\newblock {\em Pages  791--799 of:} {\em Advances in Neural Information
  Processing Systems}.

\bibitem[\protect\citename{Spring \& Shrivastava, }2017]{spring2017new}
Spring, Ryan, \& Shrivastava, Anshumali. 2017.
\newblock A New Unbiased and Efficient Class of LSH-Based Samplers and
  Estimators for Partition Function Computation in Log-Linear Models.
\newblock {\em arXiv preprint arXiv:1703.05160}.

\bibitem[\protect\citename{Wang \& Blei, }2012]{wang2012split}
Wang, Chong, \& Blei, David~M. 2012.
\newblock A split-merge MCMC algorithm for the hierarchical Dirichlet process.
\newblock {\em arXiv preprint arXiv:1201.1657}.

\bibitem[\protect\citename{Wang \& Russell, }2015]{wang2015smart}
Wang, Wei, \& Russell, Stuart~J. 2015.
\newblock A Smart-Dumb/Dumb-Smart Algorithm for Efficient Split-Merge MCMC.
\newblock {\em Pages  902--911 of:} {\em UAI}.

\end{thebibliography}
	\bibliographystyle{authordate1}

\appendix
\section{Naive Hashing based Proposal Design}
	In this section, we provide the derivations of the proposal distributions of the naive LSS based Proposal Design.
	
	\subsection{Naive Smart-split/Dump Merge}
	The probability $q(x'|x)$ of the LSH Smart-split move is:
	\[
	\begin{split}
	q(x'|x) &= \left(\frac{1}{2}\right)^{|C_u| + |C_v| - 2} \\&\sum_{u}^{C_u} \sum_{v}^{C_v} \left( \frac{1}{n} \left(1-\left(1-Pr(-u,v)^K\right)^L \right) \frac{|C_v \cap S_{-u}|}{|S_{-u}|} \right).
	\end{split}
	\]
	
	For the LSH Smart-split move, $q(x'|x)$ denotes that: given the state $x$, what is the probability that we can go from $x$ to $x'$.
	In this setting, state $x'$ contains components $C_u$ and $C_v$, and state $x$ contains the original component $C$.
	So, under our LSH split proposal design, every combination of $ u\in C_u$ and $v \in C_v$ can lead state $x$ goes to state $x'$.
	So, we sum over all possible combinations of $u$ and $v$ in our probability.
	Once we choose $u$ and $v$, then splitting their components $C$ into $C_u$ and $C_v$ has the probability of $\left(\frac{1}{2}\right)^{|C_u| + |C_v| - 2}$. We multiply them all together and yield the desired expression.
	
	Inside the summation, for each particular $u$ and $v$, we have the probability of sampling them under our proposal: \[\left( \frac{1}{n} \left(1-\left(1-Pr(-u,v)^K\right)^L \right) \frac{|C_v \cap S_{-u}|}{|S_{-u}|} \right)\]. The above expression is obtained by combining the fact that $1/n$ is the probability of choosing $u$. $(1-(1-Pr(-u,v)^K)^L)$ is the probability of having $v$ in the buckets probed. $\frac{|C_v \cap S_{-u}|}{|S_{-u}|}$ is the probability of getting $v \in C_v$ by randomly sampling the bucket. Multiply these three expression together will lead to the desire expression: $\left( \frac{1}{n} \left(1-\left(1-Pr(-u,v)^K\right)^L \right) \frac{|C_v \cap S_{-u}|}{|S_{-u}|} \right)$.
	
	The corresponding inverse move (Dump Merge) probability $q(x|x')$ is as follow:
	\[
	q(x|x') = \frac{2}{M_{x'}(M_{x'}-1)}.
	\]
	
	The inverse move of the LSH Smart-split move is a dumb move: given the state $x'$ which contains $C_u$ and $C_v$, what is the probability that we can combine $C_u$ and $C_v$ to generate $C$ (state $x$).
	Under dumb merge, we have $\frac{1}{M_{x'}}$ to choose one of $C_u$ or $C_v$, and probability $\frac{1}{M_{x'}-1}$ to choose another. Then we add a factor of two and multiplier of these two probabilities to obtain the corresponding probability. 
	
	\subsection{Naive LSH Smart-merge/Dumb-split}
	The probability of the LSH merge move $q(x'|x)$ is:
	\[
	\begin{split}
	&q(x'|x) = \sum_{u}^{C_u} \sum_{v}^{C_v} \left( \frac{1}{n} \left(1-\left(1-Pr(u,v)^K\right)^L \right) \frac{|C_v \cap S_{u}|}{|S_{u}|} \right).
	\end{split}
	\]
	
	The probability of the LSH Smart-merge move is given the state $x$, what is the probability that we can go from $x$ to $x'$.
	In our setting, state $x'$ contains the merged component $C$, and state $x$ contains the original components $C_u$ and $C_v$.
	So, under our LSH merge proposal design.
	Every combination of $ u\in C_u$ and $v \in C_v$ can lead to the state $x'$ from state $x$. The analysis of this probability is the same of LSH merge proposal distribution that introduced before. Once we choose $u \in C_u$, and $v \in C_v$, we have probability $1$ to merge $C_u$ and $C_v$ together to get $C$ under our proposal. So we multiply them all together and yield the desired expression.
	
	The corresponding inverse (Dump Split) probability $q(x|x')$ is:
	\[
	q(x|x') = \frac{1}{M_{x'}}(\frac{1}{2})^{|C_u| + |C_v|}.
	\]
	
	The inverse move of the LSH Smart-merge move is a dumb split: given the state $x'$ which contains the combined component $C$, what is the probability that we can split $C$ to $C_u$ and $C_v$ (state $x$).
	Under dumb split, we have $\frac{1}{M_{x'}}$ to choose component $C$. Then we do a random split, so we have probability $(\frac{1}{2})^{|C_u| + |C_v|}$ to split $C$ into $C_u$ and $C_v$. We multiply these two expressions together and yield the desired expression.
	
	\section{Minwise Hashing based Proposal Design}
	
	In this section, we provide the derivations of the proposal distributions of the Minhash LSS based proposal design.
	
	\subsection{MinHash Smart-split/Dumb-merge}
	Given a new state $x'$, and the corresponding old state $x$. The corresponding probability of the inverse move is calculated as:
	\begin{equation}
	q(x|x') = \frac{2}{M_{x'}(M_{x'}-1)}.
	\end{equation}
	The inverse move of the smart-split move is a dumb move: given the state $x'$ which contains $C_u \cap S_u$ and $C_u - S_u$, what is the probability that we can combine $C_u \cap S_u$ and $C_u - S_u$ to generate $C_u$ (state $x$).
	Under dumb merge, we have $\frac{1}{M_{x'}}$ to choose one of $C_u \cap S_u$ or $C_u - S_u$, and probability $\frac{1}{M_{x'}-1}$ to choose the other one. Then we add a factor of two and multiplier of these two probabilities to obtain the corresponding probability. 
	
	\subsection{MinHash Smart-merge/Dumb-split}
	
	Given a new state $x'$, and the corresponding old state $x$. The probability of the merge move $q(x'|x)$ and the corresponding inverse probability $q(x|x')$ is calculated as:
	\begin{equation}
	q(x'|x) = \frac{1}{M_{x}}\frac{\sum_{j}^{2D} \min\{{u}_j, {v}_j\}}{\sum_{j}^{2D} \max\{{u}_j, {v}_j\}} \frac{1}{|S_{u}|}.
	\end{equation}
	
	\begin{equation}
	q(x|x') = \frac{1}{M_{x'}}(\frac{1}{2})^{|C_u| + |C_v|}.
	\end{equation}
	
	Under our LSH merge proposal design, we have the probability $\frac{1}{M_{x}}$ to choose the component $C_u$.
	Then, we have the probability $ \frac{\sum_{i}^{2D} \min\{{u}_i, {v}_i\}}{\sum_{i}^{2D} \max\{{u}_i, {v}_i\}} \frac{1}{S_{u}}$ to pick cluster center $v$ in the hash table. Here, $ \frac{\sum_{i}^{2D} \min\{{u}_i, {v}_i\}}{\sum_{i}^{2D} \max\{{u}_i, {v}_i\}}$ is the probability that two centers $u$ and $v$ hashed into the same bucket, and $\frac{1}{S_{u}}$ is the probability to pick $v$ from the bucket. We multiply these two expressions together and yield the desired expression.
	
	The inverse move of the LSH Smart-merge move is a dumb split: given the state $x'$ which contains the combined component $C$, what is the probability that we can split $C$ to $C_u$ and $C_v$ (state $x$).
	Under dumb split, we have $\frac{1}{M_{x'}}$ to choose component $C$. Then we do a random split, so we have probability $(\frac{1}{2})^{|C_u| + |C_v|}$ to split $C$ into $C_u$ and $C_v$. We multiply these two expressions together and yield the desired expression.
\end{document}